\documentclass[letterpaper, 10 pt, conference]{ieeeconf}  % Comment this line out if you need a4paper

\IEEEoverridecommandlockouts                              % This command is only needed if 
\makeatletter
\let\NAT@parse\undefined
\makeatother

\usepackage{graphicx} % for pdf, bitmapped graphics files%\usepackage{epsfig} % for postscript graphics files
\usepackage[hidelinks]{hyperref}
\usepackage{amsfonts}
\usepackage{amsmath} % assumes amsmath package installed
\usepackage{amssymb}  % assumes amsmath package installed
\usepackage{url}

\usepackage{booktabs} % Pour des tableaux plus esthétiques
\usepackage{array} % Pour les options de tableau supplémentaires
\usepackage{color}
\usepackage{multirow}
\usepackage{booktabs}
\usepackage{cite}
\usepackage{threeparttable}

\title{\LARGE \bf
FRASA: An End-to-End Reinforcement Learning Agent for Fall Recovery and Stand Up of Humanoid Robots
}

\author{Clément Gaspard$^{1*}$, Marc Duclusaud$^{1*}$, Grégoire Passault$^{1}$, Mélodie Daniel$^{1}$,  Olivier Ly$^{1}$% <-this % stops a space
\thanks{$^{1}$Univ. Bordeaux, CNRS, LaBRI, UMR 5800, 33400 Talence, France. Corresponding author: Clément Gaspard, e-mail: \texttt{clement.gaspard@u-bordeaux.fr}.
\newline This study has received financial support from the French government in the framework of the France 2030 program, Initiative of Excellence (IdEx) University of Bordeaux / RRI ROBSYS and from the ENS Rennes.
\newline $^{*}$ These authors contributed equally.}}

\begin{document}

% \begin{figure*}[b]
%     \centering
%     \includegraphics[width=\linewidth]{img/CatchEye.pdf}
%     \caption{Recovery agent (AJOUT DES MEMES ETAPES DU DOS PREVU)}
%     \label{fig:enter-label}
% \end{figure*}

\maketitle
\thispagestyle{empty}
\pagestyle{empty}

% \vspace*{-1\baselineskip}
%%%%%%%%%%%%%%%%%%%%%%%%%%%%%%%%%%%%%%%%%%%%%%%%%%%%%%%%%%%%%%%%%%%%%%%%%%%%%%%%
\begin{abstract}

% In recent years, the field of humanoid robotics has witnessed significant advancements, particularly in developing robots capable of autonomously performing complex tasks in dynamic environments. One of the primary challenges remains ensuring stability and effective fall recovery, crucial for humanoids due to their inherently unstable structure. This paper presents FRASA, an end-to-end Reinforcement Learning approach for adaptive fall prevention and recovery of humanoid robots, integrating both fall recovery and disturbance rejection into a single agent.

% FRASA leverages an effective reward structure to train the agent in a simulated environment, achieving reduced training time and enhanced robustness. The agent uses servo positions and IMU data as inputs and controls the robot’s joints directly to maintain stability and recover from falls. Experiments conducted on the Sigmaban platform, a 70cm-high humanoid robot, demonstrate the efficacy of the approach in real-world scenarios. The results indicate that FRASA significantly outperforms traditional key-frame-based methods, providing a more adaptive and versatile solution for humanoid recovery.

Humanoid robotics faces significant challenges in achieving stable locomotion and recovering from falls in dynamic environments. Traditional methods, such as Model Predictive Control (MPC) and Key Frame Based (KFB) routines, either require extensive fine-tuning or lack real-time adaptability. This paper introduces FRASA, a Deep Reinforcement Learning (DRL) agent that integrates fall recovery and stand up strategies into a unified framework. Leveraging the Cross-Q algorithm, FRASA significantly reduces training time and offers a versatile recovery strategy that adapts to unpredictable disturbances. Comparative tests on Sigmaban humanoid robots demonstrate FRASA superior performance against the KFB method deployed in the RoboCup 2023 by the Rhoban Team, world champion of the KidSize League.

\end{abstract}

%%%%%%%%%%%%%%%%%%%%%%%%%%%%%%%%%%%%%%%%%%%%%%%%%%%%%%%%%%%%%%%%%%%%%%%%%%%%%%%%
\section{Introduction}\label{sec:intro} % MARC 

Humanoid robotics has made significant strides in recent years, driven by advancements in both hardware and machine learning. One of the key challenges in this field is developing robots that can autonomously perform complex tasks, including locomotion, in dynamic and unpredictable environments~\cite{Saeedvand_Jafari_Aghdasi_Baltes_2019}. This is the case in several humanoid robotics competitions or challenges, such as RoboCup~\cite{kitano1997robocup}, the DARPA Robotics Challenge~\cite{darpa}, and the recent Avatar XPRIZE~\cite{xprize_avatar}. Reinforcement Learning (RL) has emerged as a powerful tool for addressing these challenges, enabling robots to learn from interactions with their environment and have more adaptive responses~\cite{peters2003reinforcement}.

% Achieving stable locomotion remains one of the greatest challenges in humanoid robotics due to the limited ground contact points and the height of the Center of Mass (CoM), making them prone to instability and falls~\cite{7337218, cai_self-protect_2023}. Thus, in the literature, fall prevention often takes precedence over recovery.

Achieving stable locomotion is one of the primary obstacles, if not the greatest, in humanoid robotics. The difficulty of this task arises primarily from the limited number of ground contact points for a humanoid, which results in an inherently unstable standing pose~\cite{7337218}. Moreover, a high Center of Mass (CoM) increases the risk of damage in the event of a fall for humanoid robots~\cite{cai_self-protect_2023}. This leads researchers to prioritize fall prevention over recovery.

A widely adopted approach in humanoid robotics to achieve stable walking and prevents falls is the use of Model Predictive Control (MPC)~\cite{katayama2023model}. This method relies on planning the trajectory of the CoM while ensuring stability criteria over a finite time horizon. The trajectory is, then, frequently updated and adjusted based on sensor feedback\cite{scianca2020mpc}.

\begin{figure}[tp]
    \centering
    \includegraphics[width=0.97\linewidth]{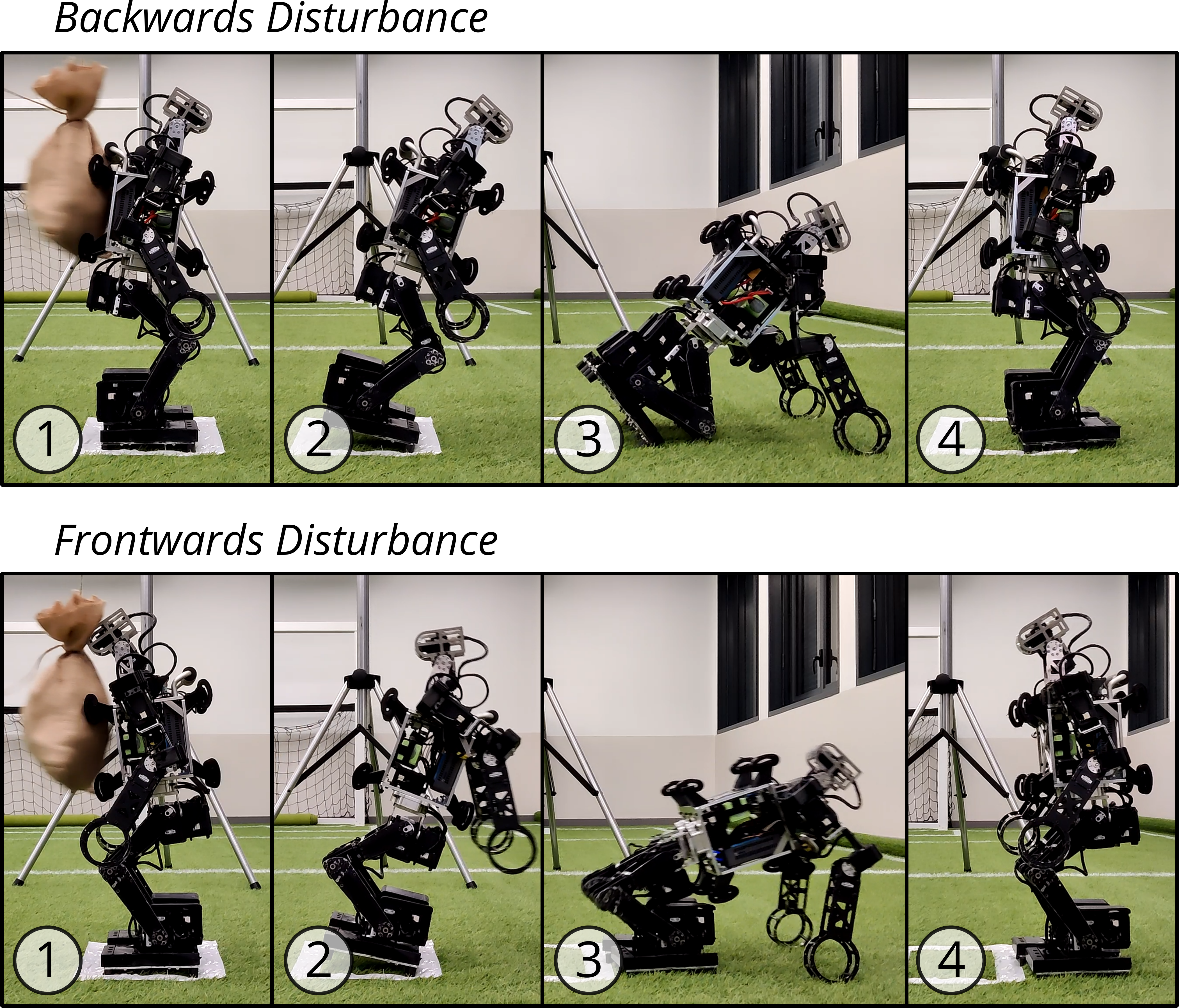}
    \caption{FRASA adaptative response to a backwards and a frontwards disturbances on the Sigmaban platform. The recovery behavior using the arms accelerates the return to a stable position while minimizing the risk of damage.}
    \label{fig:catcheye}
    \vspace*{-1.8\baselineskip}
\end{figure}

When significant deviations from the planned CoM trajectory occur, such as due to an external disturbance, three solutions are generally presented in push recovery problems: modifying the center of pressure using ankle joints, generating angular momentum with hip joints, or taking a step~\cite{shafiee2016push}~\cite{MissuraCP}. One usually uses the capture point to determine the good strategy ~\cite{capture}. The CP is a specific measure derived from the linear inverted pendulum model~\cite{kajita20013d}, corresponding to the location where a robot should step to halt its linear momentum and eventually come to a stop. While effective, this method is constrained by arbitrary adjustment choices, strong linearity assumptions, and significant computational complexity.

More recently, other approaches involving training Deep Reinforcement Learning (DRL) agents have emerged to address push recovery problems~\cite{aslan2023Development}~\cite{kim2019Push}~\cite{vanmarum2024revisiting}. These methods address the drawbacks of MPC-based approaches but typically require extensive training times. Moreover, these approaches do not offer recovery solutions after a fall. 

%Modif Clem
This paper does not focus on the problem of push recovery but rather on \textbf{fall recovery}, which involves creating new points of contact with the ground as a strategy to regain stability in minimal time. This approach is crucial when external forces are too strong on the robot to prevent a fall.
In robotics competitions~\cite{kitano1997robocup}\cite{darpa}\cite{xprize_avatar}, \textbf{fall recovery} is usually handled by allowing the fall to complete before initiating a \textbf{stand-up} routine, which is often based on key frame animations~\cite{stuckler2006getting}.
%Modif Clem
% The \textbf{fall recovery} approach is usually used in robotics competitions~\cite{kitano1997robocup}~\cite{darpa}~\cite{xprize_avatar} is to wait for the fall to end before starting a \textbf{stand up} routine, generally based on key frame animations~\cite{stuckler2006getting}. 
This method is simple to implement but lacks versatility, requires extensive fine-tuning, and must be frequently adjusted due to motor wear and loss of precision.

Another approach proposes using genetic algorithms to optimize the parameters of stand up splines~\cite{marc}. This method offers the advantage of generalizing a stand up routine across different humanoid architectures. However, it remains an open-loop animation, limiting real-time adaptability to unexpected disturbances.

Recent RL-based studies tackle the stand up challenge. A Q-learning approach in \cite{jeong} incorporates environment feedback but relies on complex problem discretization and lacks fall recovery. Another method trains DRL agents to learn stand up motions~\cite{haarnoja_learning_2024}, but it still depends on expert key frames and requires approximately one day of training.

In this paper, we propose a Fall Recovery And Stand up Agent (\textbf{FRASA}) that integrates both fall recovery and stand up strategies into a single agent, aiming to resume walking as fast as possible after a disturbance. The adaptative response allowed by FRASA in the case of backwards and frontwards disturbances is presented in Fig.~\ref{fig:catcheye}. 

The key contributions of FRASA are:

\begin{itemize}
    \item \textbf{Unified task handling}: The proposed reward function enables the DRL agent to efficiently address both fall recovery and stand up tasks within a unified framework.
    \item \textbf{Reduced training time}: By leveraging the Cross-Q algorithm, FRASA achieves effective training in significantly less time compared to existing RL approaches~\cite{haarnoja_learning_2024}.
    \item \textbf{Versatile and adaptative recovery strategy}: The use of a DRL agent allows adaptation to unpredictable real-world feedback and recovery from various postures without the need for expert tuning.
\end{itemize}

Tests are conducted on Sigmaban humanoid robots~\cite{kidsize_site} to compare FRASA with a Key Frame Based (KFB) approach \cite{stuckler2006getting}. The KFB method used for comparison was deployed in the RoboCup 2023 by the Rhoban Team, world champion of the KidSize League. Real robots experiments are presented in the following video\footnote{\href{https://youtu.be/VRgBEy0yhfU}{https://youtu.be/NL65XW0O0mk}} and the open-source code for this project is available on our GitHub repository\footnote{\href{https://github.com/Rhoban/frasa}{https://github.com/Rhoban/frasa}} for further use and contributions.

\section{Problem Statement}\label{sec:pb_statement} % MARC

% Some humanoid robots must operate in highly unpredictable environments where interactions with other robots or objects can cause disturbances so severe that falls become inevitable.

%This limitation arises from the lack of versatility of these methods, which need to be initiated from precise positions to ensure the success of the movement. 

The problem we aim to solve is twofold: first, the ability to stand up from a prone or supine position, and second, to recover when a fall is about to occur. The fall recovery problem necessitates creating new contact points with the ground in cases of strong or repeated disturbances. In both cases, the objective is to resume movement, specifically walking, as quickly and smoothly as possible.

Let us note that, the humanoid platform's design allows it to naturally roll into prone or supine positions after a fall. In addition, reaching these positions after a fall is generally a feasible task for humanoids. Consequently, the stand up problem can be restricted to these two fall configurations.

\section{Background}\label{sec:background} % CLEMENT 
% Bagage necessaire à la compréhension de l'article
% Bagage RL

\subsection{RL and DRL}

RL is a machine learning field where an agent learns to make decisions by interacting with its environment to maximize cumulative rewards, typically modeled as a Markov Decision Process (MDP). It is represented by the tuple $(\mathcal{S}, \mathcal{A}, P, R)$ \cite{puterman2014markov}, where $\mathcal{S}$ is the state space, $\mathcal{A}$ is the action space, $P$ is the transition function, and $R$ is the reward function. The agent's goal is to find an optimal policy that maximizes the expected episode reward, represented as a discounted sum of future rewards.

DRL uses deep neural networks to approximate the action-value function and the policy, particularly through actor-critic architectures \cite{tuomas2018soft}. The critic estimates the value of state-action pairs using temporal difference learning, while the actor updates the policy to maximize this value, enhancing exploration through added noise. Recent DRL advancements have improved policy and action-value function approximations, enabling effective solutions for complex decision-making tasks in high-dimensional environments \cite{tuomas2018soft} \cite{fujimoto2018addressing}.

\subsection{CrossQ Algorithm}

% Training DRL agents can be time-consuming due to the complexity of the environments. A recent algorithm, CrossQ, has been introduced to improve sampling efficiency and reduce training time \cite{bhatt2024crossq}. It reduces training time by removing target networks, normalizing batches and using wider critic layers. CrossQ can be used in addition to a classical actor-critic architecture that uses two critic networks, such as the Soft Actor-Critic (SAC) architecture \cite{tuomas2018soft}.

Training DRL agents is often time-consuming due to complex environments. A recent algorithm, CrossQ, improves sampling efficiency and reduces training time by removing target networks, normalizing batches, and using wider critic layers~\cite{bhatt2024crossq}. It can be integrated with traditional actor-critic architectures, like Soft Actor-Critic (SAC)~\cite{tuomas2018soft}.

% Traditionally, target networks are used to stabilize the training but slow it down due to delayed updates \cite{fujimoto2018addressing} \cite{sinigaglia2024exploiting}. Thus, eliminating target networks simplifies the architecture and reduces the computational cost. In addition, CrossQ stabilizes training by normalizing the inputs of each layer, thus reducing the offset of internal covariates. Furthermore, CrossQ uses larger critic networks to increase the efficiency of the learned policies.

 Empirical results show that CrossQ matches or surpasses state-of-the-art algorithms in sample efficiency while significantly reducing computational costs \cite{bhatt2024crossq}, making it a promising approach for continuous control tasks in DRL.

\section{Method}\label{sec:method} %(ou Approach) CLEMENT

% Modif Clément
In this paper, we present FRASA: an end-to-end DRL approach for humanoid robot autonomous stand up and fall recovery from external disturbances. The agent is trained in a full-body physics simulator \cite{6386109}, incorporating a robot's detailed collision model. In simulation, the robot's degrees of freedom (DoFs) are position controlled to reflect servomotors behavior. The core contribution of FRASA lies in formulating an environment that enables efficient learning with actor-critic algorithms, such as SAC and, more recently, CrossQ.

The RL environment leverages the inherent symmetrical design of the robot to simplify the learning process. To this end, only 5 of the robot's DoFs are observed and controlled, as shown in Fig.~\ref{fig:dof}. This selection is strategic, as these are the primary DoFs involved in the plane of movement $(x,z)$. 

The target posture for the recovery movement is the initial stance associated with walking, noted as neutral pose in Fig~\ref{fig:dof}. This target pose is defined by the angles of the controlled joints $q_{\text{target}}$ and the robot's trunk pitch $\theta_{\text{target}}$ :
\begin{equation}
\label{equ:target}
    \psi_{\text{target}} = [q_{\text{target}}, \theta_{\text{target}}]
\end{equation}

By adjusting the target posture, it is possible to modify the subsequent movement after recovery, enhancing the versatility of the solution.

%Modif Clément

%This choice is also made because these are the only DOFs in the plane of movement. \Todo A voir si le fait qu'on soit dans un plan du mouvement n'est pas plutôt quelque chose que l'on a choisi et donc que ça reste une simplification \Done 

\subsection{Environment initialization}

At the start of each episode, the robot's joint angles $q$ and trunk pitch $\theta$ are set to random values within their physical limits. The robot is then released above the ground, simulating various fallen states. The physical simulation is then stepped using current joints configuration as targets, until stability is reached, at which point the learning process begins.

% \newpage

% \par
% ~
\vspace{-1.em}

\begin{figure}[!ht]
    \centering
    \includegraphics[width=.35\textwidth]{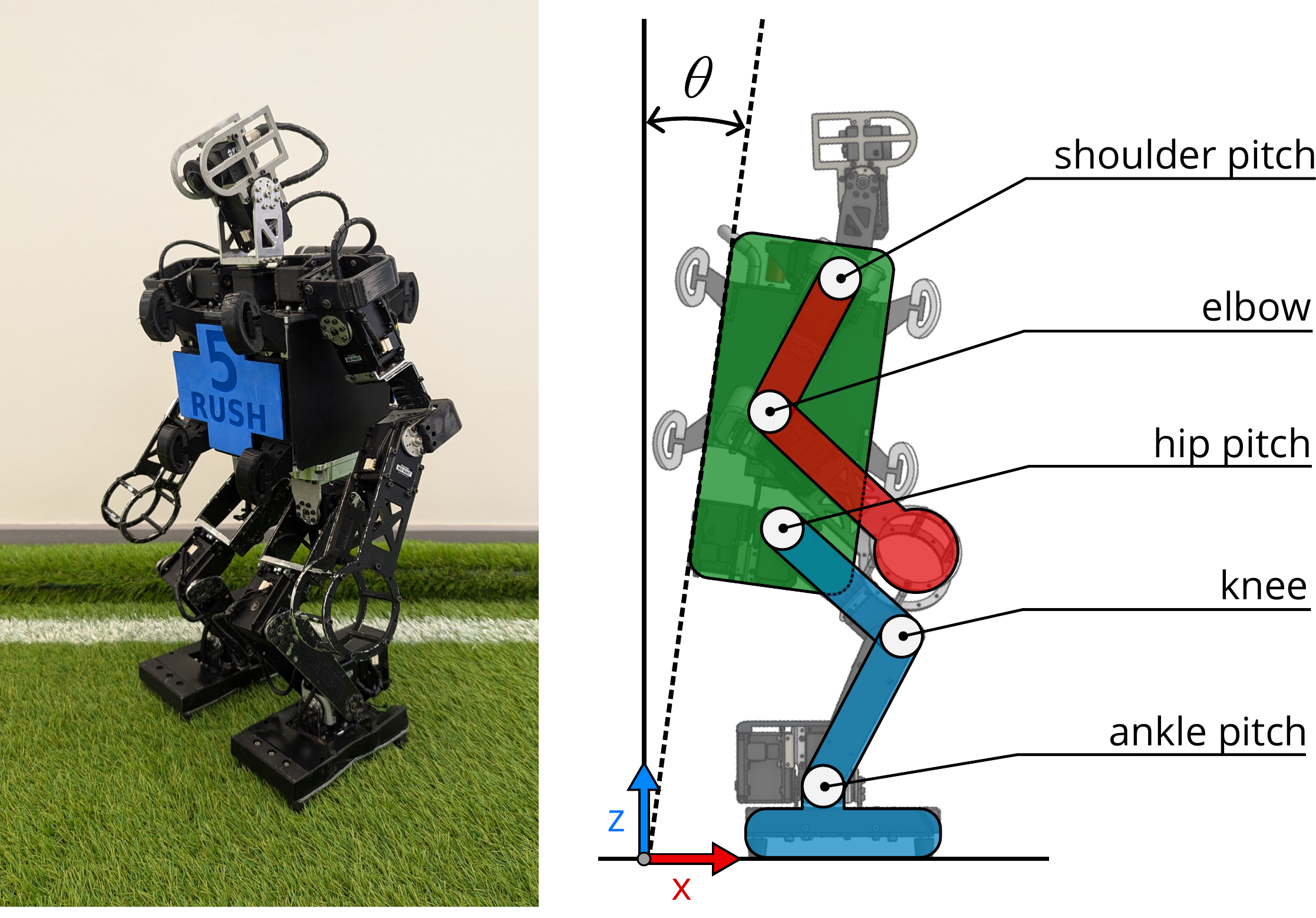}
    \vspace*{-.5\baselineskip}
    \caption{\textbf{Left}: target posture for the recovery movement. \textbf{Right}: pose vector $\psi_{target}$ components, including the trunk pitch $\theta$ and the 5 explicited DoFs.}
    \vspace*{-.8\baselineskip}
    \label{fig:dof}
\end{figure}

This approach prevents overfitting to specific starting positions and promotes exploration of a wide range of configurations, enabling the development of a general policy effective across diverse scenarios. Furthermore, the robot's state is occasionally initialized close to the neutral pose presented in Fig.~\ref{fig:dof} with a certain probability, $p_s$. This strategy provides occasional large rewards, reinforcing successful standing behaviors and thus speeding up the learning process.

\subsection{Action Space} As this is an end-to-end model, the action space is the control commands that can be symmetrically applied to the joints on both sides of the robot. The action space is defined as a variation of the desired joints configuration :
\begin{equation}
\label{equ:action_space}
a_t = \dot{q}^{\text{desired}}_t
\end{equation}

During each simulation time step $t$, the desired joint angles \( q^{\text{desired}}_t \) transition from their initial values to \( q_{t+1}^{\text{desired}} = q^{\text{desired}}_t + \Delta t \times \dot{q}^{\text{desired}}_t \), reflecting the incremental control applied by the agent.
\subsection{State Space} 
The state space captures the robot's posture and movements at each time step. It is designed as 
\begin{equation}
\label{equ:state_space}
s_t = \begin{bmatrix} q_t, \Dot{q_t},  q^{\text{desired}}_t, \theta_t, \dot{\theta_t}, a_{t-1}, a_{t-2}, \ldots, a_{t-k} \end{bmatrix},
\end{equation}
where \( q_t \) and \( \Dot{q_t} \) are the joint angles and velocities of the 5 DoFs defined in Fig.~\ref{fig:dof}. They are determined as the average values of the corresponding actuators on both sides of the robot for each degree of freedom. \( \theta_t \) and \( \dot{\theta_t} \) are respectively the trunk pitch and its rate of change.
Given \eqref{equ:action_space}, $a_{t-1}, a_{t-2}, \ldots, a_{t-k}$ represent the target joint position update rates issued by the agent over the last $k$ time steps.

\subsection{Reward}
The reward function is designed to encourage the robot to achieve and maintain the target upright posture (cf. Fig.~\ref{fig:dof}). Our reward function is designed as followed with the different components defined in Table~\ref{tab:reward_function} : 
\begin{equation}
R = R_{\text{state}} + R_{\text{variation}} + R_{\text{collision}}
\end{equation}

In $R_{\text{state}}$, the current state $\psi_t = \begin{bmatrix} q_t, \theta_t \end{bmatrix}$ is compared to the target state $\psi_{\text{target}}$ defined in \eqref{equ:target}. A Gaussian function shapes $R_{\text{state}}$, providing higher rewards as the robot's state approaches the desired target. The parameter $w_1$ shapes the reward, determining how sharply it decreases as the robot's state deviates from the desired state, thus encouraging the robot to achieve and maintain the desired posture. $R_{\text{variation}}$ penalizes high control velocities and large variations in actions, promoting smoother movements. $R_{\text{collision}}$ discourages self-collisions, determined using the physics simulator.

\vspace*{-.3\baselineskip}

\begin{table}[!ht]
\centering
\caption{Components of the Reward Function}
\vspace*{-1em}
\resizebox{0.95\columnwidth}{!}{%
\begin{tabular}{l p{4.5cm}}
\toprule
\textbf{Reward components} & \textbf{Formulas} \\ \midrule
Desired state proximity reward & \( R_{\text{state}} = \exp(-w_1 \times \| \psi_t - \psi_{\text{target}} \|^2) \) \\ \midrule
Action variation reward & \( R_{\text{variation}} = w_2 \times \exp(-\| a_t - a_{t-1} \|) \) \\ \midrule
No self-collision reward & \( R_{\text{collision}} = w_3 \times \exp(-\text{self\_collisions}) \) \\ \bottomrule
\end{tabular}%
}
\label{tab:reward_function}
\vspace*{-0.7\baselineskip}
\end{table}

The weights \( 0 \le w_2, w_3 \ll 1 \) are small penalizations intended to smooth learning and prevent the robot from adopting mechanically feasible but undesirable strategies.

When the robot reaches the desired position, the reward is approximately 1 thanks to $R_{\text{state}}$, providing a clear unit of measure for the agent's performance in terms of maintaining upright posture over time.

% \begin{table}[h]
% \centering
% \caption{Components of the Reward Function in the StandupEnv Environment}
% \resizebox{\columnwidth}{!}{%
% \begin{tabular}{l p{4.5cm}}
% \toprule
% \textbf{Component} & \textbf{Description} \\ \midrule
% \multirow{2}{*}{\textbf{Desired State Proximity Reward}} & \( R_{ds} = \exp(-w_1 \times \| z_t - z_{\text{desired}} \|^2) \) \\ 
% & Encourages the robot to achieve and maintain the desired state. \\ \midrule
% \multirow{2}{*}{\textbf{Action Variation Rewards}} & \( R_{av} = w_2 \times \exp(-\| a_t - a_{t-1} \|) \) \\ 
% & Penalizes high control velocities and large variations in actions to promote smooth movements. \\ \midrule
% \multirow{2}{*}{\textbf{No Self-Collision Reward}} & \( R_{sc} = w_3 \times \exp(-\text{self\_collisions}) \) \\ 
% & Provides a small reward to discourage self-collisions. The self-collisions are determined using the physics simulator. \\ \bottomrule
% \end{tabular}%
% }
% \label{tab:reward_function}
% \end{table}

\subsection{Episode termination and truncation} 
An episode termination occurs when the robot reaches an unsafe or irreversible state:
\begin{itemize}
    \item Trunk pitch ($\theta$) exceeds a predefined threshold, indicating the robot is upside down.
    \item Trunk pitch velocity ($\dot{\theta}$) surpasses a critical threshold, suggesting very violent displacements of the pelvis.
\end{itemize}

An episode is truncated if its duration exceed a predefined maximum duration. This allows for an increase in the diversity of encountered states.

\subsection{Domain Randomization} 

To enhance the robustness and generalization of the learned policy, we employ extensive domain randomization \cite{tobin2017domain} in our simulation environment. The following parameters are randomized:

\begin{itemize}
    \item Angular errors, at the beginning of each episode, to simulate the fact that parts could be slightly deformed or zero positions slightly incorrect.
    \item The mass and center of mass positions of the robot's body parts to ensure that the agent could handle variations in its own weight distribution.
    \item The friction coefficients of the contact surfaces to account for different types of ground interactions the robot might encounter.
    \item The battery voltage supply, affecting the proportional gain and maximum available torque of the actuators, simulating fluctuations in power that might occur in real-world conditions.
    \item The friction parameters of the actuators themselves, such as dry friction and damping, to reflect the wear and variability of mechanical components over time.
\end{itemize}

Table~\ref{table:randomization} provides the specific parameters and their ranges used for domain randomization. The randomization ensures that the policy learned by the robot is robust and adaptable to a wide range of real-world scenarios.

The DRL algorithm operates in an end-to-end manner, where raw state inputs are mapped directly to action outputs without any intermediate processing steps.

% Overview of the proposed method
% Insister sur le caracère end-to-end
% On considère que le robot est symétrique
% Condition(s) de finalisation de l'épisode
% Explicaiton state/action space, Reward, termination (Reward sous forme de Gaussienne pour guider l'apprentissage - position cible idéale mais plus on est proche mieux c'est)

\section{Experiments} % MARC

To validate the performance of FRASA in a real environment, we deploys it on the Sigmaban robotic platform.

\subsection{Robotic Platform}

The Sigmaban is a 70 cm-tall humanoid robot weighing 7.5 kg~\cite{allali2024rhoban}. Its limbs are milled from aluminum and it is actuated by MX-106 and MX-64 Dynamixel servos with proportional position control. The robot features a camera, an Inertial Measurement Unit (IMU), and eight foot pressure sensors using strain gauge~\cite{7363498}. On-board computations are handled by an Intel NUC (Core i5-7260U, 8 GB RAM) running Ubuntu 22.04 and a custom STM32-based board.

\subsection{Physical Simulation and Modeling}

The physics engine used for training is MuJoCo\cite{6386109}. A precise CAD model of the robot is exported to MuJoCo\cite{onshape-to-robot}, ensuring accurate geometry, mass distribution, and joint limits, closely matching the real robot.

\begin{figure}[h!]
    \centering
    \includegraphics[width=.3\textwidth]{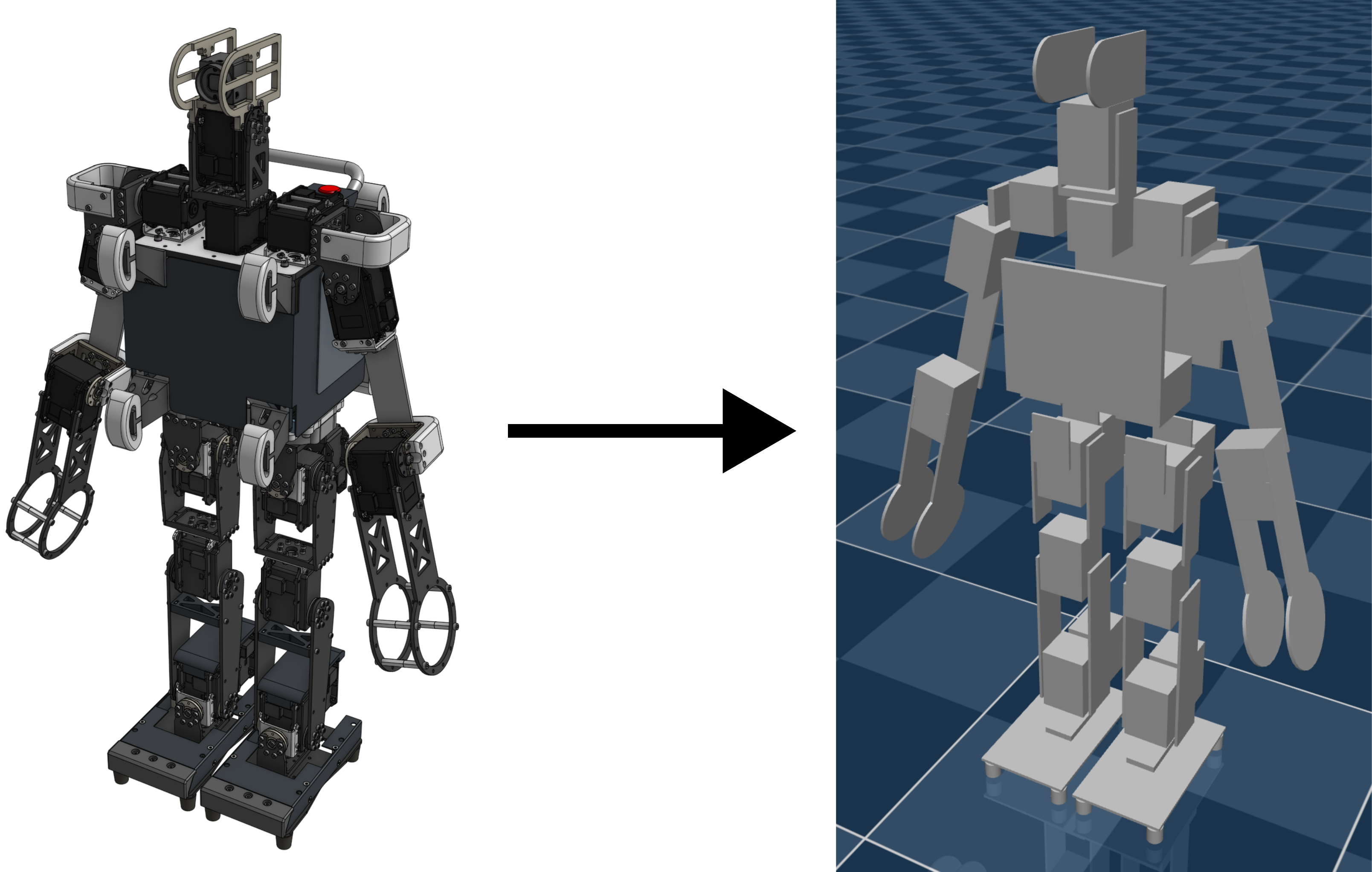}
    \caption{Approximation of the robot using primitive shapes dedicated to simulating collisions}
    \label{fig:shapes}
    \vspace*{-0.5\baselineskip}
\end{figure}

It is worth noting that in order to manage contacts effectively in the simulated environment, every part of the robot is approximated using basic shapes such as cylinders, cuboids, or combinations thereof. In total, the robot is approximated by 51 boxes and 14 cylinders, the arrangement of which is shown in Fig.~\ref{fig:shapes}.

To ensure a realistic behavior of the actuators in simulation, an identification of their friction parameters, such as dry friction and damping, is performed using the CMA-ES genetic algorithm on logs generated from a test bench. The method used is the same as in~\cite{fabre2017dynaban}.

\subsection{Training and Inference} % Clément

The agent is trained using CrossQ \cite{bhatt2024crossq} applied to SAC DRL algorithm \cite{tuomas2018soft}, implemented in Stable Baselines X, a GPU-accelerated version of Stable Baselines 3 \cite{stable-baselines3}. Utilizing the enhanced computational efficiency of this implementation, the training process is remarkably fast, resulting in a fully trained network within only 13 to 37 minutes, with a total of 235,000 to 575,000 timesteps. The training is conducted on a PC equipped with an Intel Core i9-12900 CPU, an NVIDIA GeForce RTX 3060 GPU, 64 GB of RAM, and an SSD.

\vspace*{-0.9\baselineskip}

\begin{figure}[!h]
    \centering
    \includegraphics[width=0.45\textwidth, trim={1cm 1.15cm 0.8cm 1.2cm}, clip]{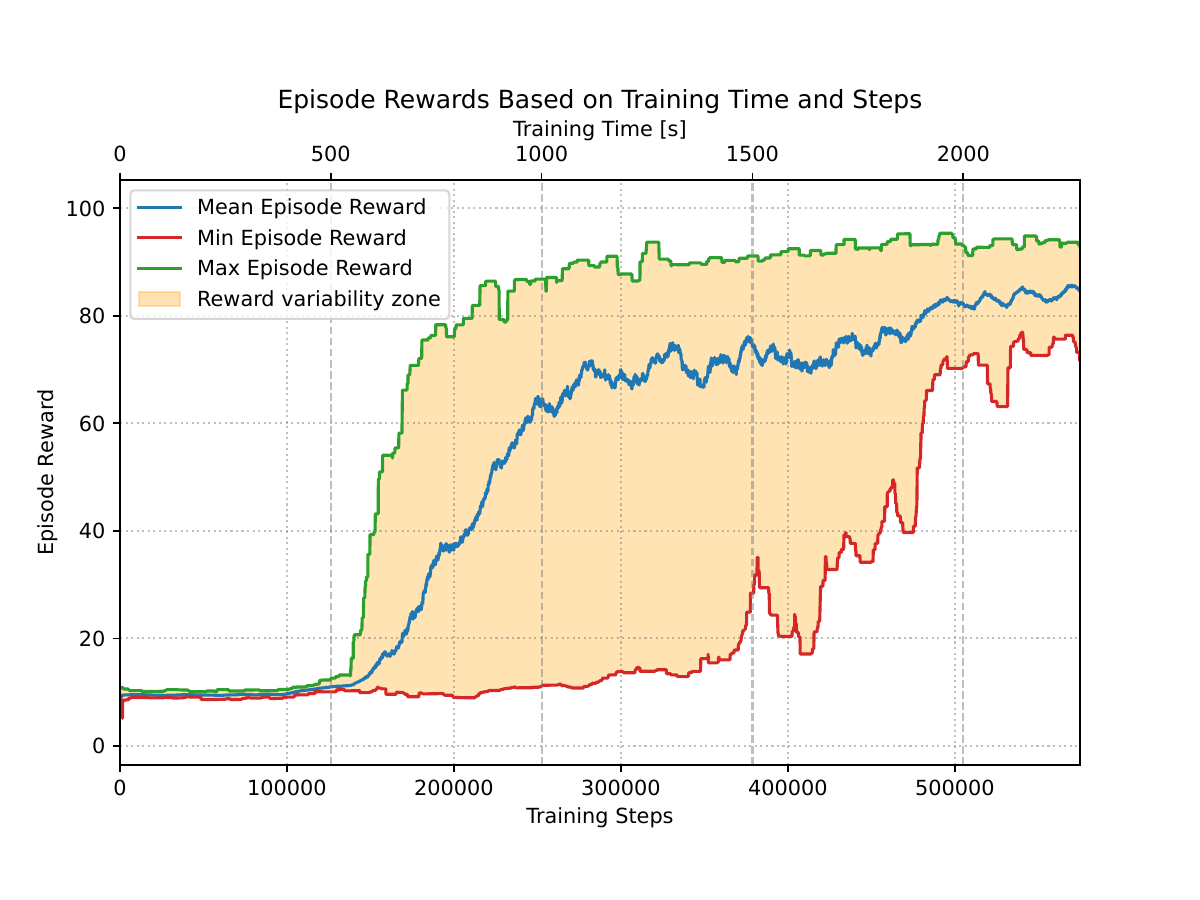}
    \vspace*{-0.8\baselineskip}
    \caption{Reward variability observed during the training of 40 distinct FRASA agents, each trained over 575,000 steps within 37 minutes using the CrossQ+SAC algorithm.}
    \label{fig:training_curve}
    \vspace*{-0.8\baselineskip}
\end{figure}

Fig.~\ref{fig:training_curve} illustrates the reward progression during the evaluation phase of 40 trained agents of FRASA. The ``reward variability zone" highlights the range of rewards experienced across the 40 evaluation sessions. 
The fastest agents learn to stand up in 13 minutes, with convergence toward a consistent reward observed across all networks after approximately 30 minutes of training.
While some agents are better trained than others, this variability is expected and does not detract from the overall efficiency and reliability of the training process.

The hyperparameters used to optimize the learning process include a learning rate set to 0.001, and a large buffer size of 1M transitions to store past experiences. Each training batch consists of 256 samples, and the training utilizes 16 parallel environments to enhance data collection efficiency. The discount factor is set to 0.99, ensuring that the agent considers long-term rewards. Training updates are applied every 4 steps, with 512 gradient descents per update, enabling the model to adjust its parameters 512 times per batch. Training starts after 100,000 steps, ensuring a diverse buffer. The policy network comprises two hidden layers containing 384 and 256 neurons respectively.

To exploit the maximum capacities of our computational power during runtime, we use the open source OpenVINO™ Runtime \cite{openvino_2023}. The mean on-board inference time to generate the actuators command for one timestep is $40\mu s$.

\subsection{Sim-to-real Transfer}
% Paragraphe sur le sim to real gap et les difficultés à surmonter

In addition to the precise modeling of the robot in MuJoCo, sensor communication delays, such as those from the IMU and the actuators, are measured on the real robot and modeled in the simulation. They are crucial to ensure a good transfer to the real robot.

Furthermore, extensive domain randomization is employed during training (see Section~\ref{sec:method}). Table~\ref{table:randomization} lists the specific parameters and their randomization ranges.

% \newpage
% \par
% ~
\vspace{-0.8em}

\begin{table}[!ht]
    \centering
    \caption{Parameters used for sim-to-real and domain randomization}
    \resizebox{0.45\textwidth}{!}{%
    \begin{tabular}{lll}
        \toprule
        \textbf{Parameter} & \textbf{Initial Value} & \textbf{Randomization Range} \\ 
        \midrule
        Floor friction coefficient & 0.875 & $\pm$0.625 \\ 
        Trunk mass & 3.28kg & $\pm$20\% \\ 
        Trunk CoM position offset & (0, 0, 0) & $\pm 5 \times 10^{-3}$ m \\ 
        Battery voltage & 15 V & [13.8, 16.8] V \\ 
        Actuators proportional gain\textbf{*} & 12.5 / 21 & Scaled by battery voltage \\ 
        Actuators abs. force bound\textbf{*} & 5 / 8 N.m & Scaled by battery voltage \\ 
        Actuators damping\textbf{*} & 0.66 / 1.7 N.m & $\pm$20\% \\ 
        Actuators friction loss\textbf{*} & 0.09 / 0.1 N.m & $\pm$20\% \\ 
        Actuator position ($q$) offset & 0 & $\pm 1.5$° \\ 
        Actuator velocities ($\dot{q}$) delay & 0.030 s & N/A \\ 
        Trunk pitch ($\theta$) delay & 0.050 s & N/A \\  
        Trunk pitch velocity ($\dot{\theta}$) delay & 0.050 s & N/A \\ 
        \bottomrule
    \end{tabular}
    }   
    \label{table:randomization}
    \par
    \vspace{1ex}
    (\textbf{*}) : MX-64 motor value / MX-106 motor value
\end{table}

%Pour enlever l'espace après le tableau
\vspace*{-0.8\baselineskip}

During the initial transfers, significant tremors appeared on the robots. The source of these tremors is a substantial discrepancy between the actuators simulated in MuJoCo and the real behavior of the motors. Specifically, the relatively simple friction model used by the simulator seems to not adequately capture the real behavior.

To solve this problem the motor gains in both the simulation and the real robot is reduced to mitigate the unforeseen dynamic effects caused by the significant gap between the simulated and real environments. Some works use downstream low-pass filters~\cite{li2024reinforcementlearningversatiledynamic}, which was unnecessary in our case but could lead to the same effects.

Increasing the inference frequency of the agent also help to reduce tremors. The agent is trained with a decision frequency of 20Hz and operates at 100Hz on the actual robot. The nature of the action space allows for this increase without necessitating significant changes.

\subsection{Stand Up Experiment}

A first experiment is conducted to compare the ability to stand up from prone and supine positions using both a KFB recovery process and FRASA. A KFB stand up method defines arbitrary key positions determined thanks to expert knowledge for the robot to pass through, interpolating motor positions between them to create a stand up trajectory \cite{stuckler2006getting}. The KFB method used in the experiments was developed for RoboCup 2023 by the Rhoban Team and demonstrated both its effectiveness and efficiency~\cite{allali2024rhoban}.

% The robot was placed lying on its back and face down, with the goal of standing up and initiating walking in place. The metric studied in this experiment is the duration between the beginning of the movement and the start of the walking phase. To initiate the start of walking, the error for each component of the state vector—comprising the 5 controlled joints and the trunk pitch—must be less than 5° for a duration of $0.5$ s.

The robot is placed on its back, face down, aiming to stand and begin walking. The metric studied is the time from movement initiation to the start of walking. To initiate walking, the error in the state vector components—comprising the 5 controlled joints and the trunk pitch—must remain below 5° for 0.5 seconds.

\subsection{Fall Recovery Experiment}

% A second experiment was implemented to compare the recovery ability after being pushed from a static standing pose using both the KFB recovery process and FRASA. To induce disturbances of varying intensities in a repeatable manner on the robot, the experimental setup depicted in Fig.~\ref{fig:setup} was constructed. It includes a pendulum mechanism designed to release a mass from various distances onto the robot. A cord connected to the pendulum enables the experimenter to retrieve the weight after impact to prevent interference with the robot during recovery. 

\begin{figure}[h!]
    \centering
    \includegraphics[width=.35\textwidth]{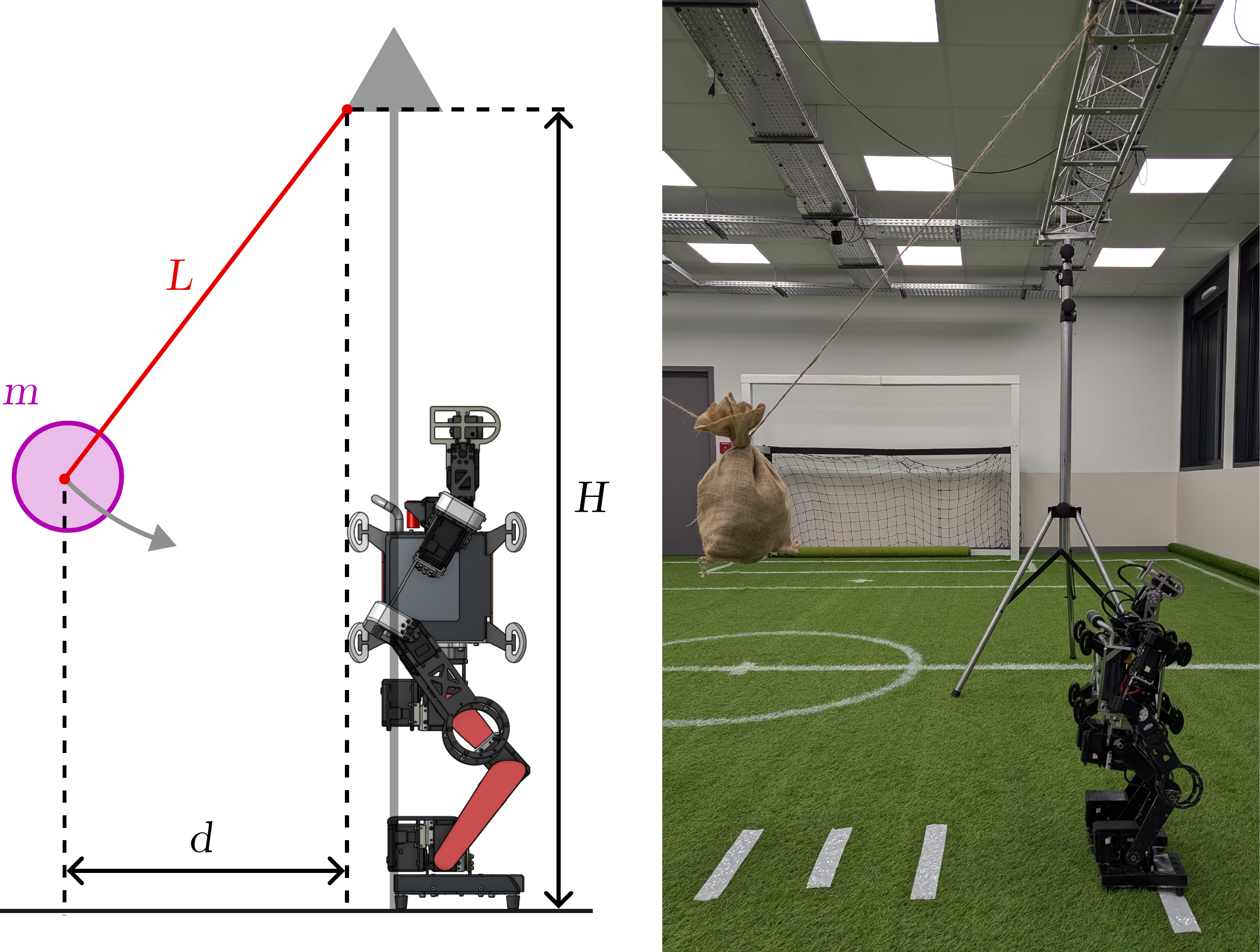}
    \vspace*{-0.5\baselineskip}
    \caption{Experimental setup for inducing repeatable disturbances}
    \label{fig:setup}
    \vspace*{-0.5\baselineskip}
\end{figure}

A second experiment compares the recovery ability of the KFB process and FRASA after being pushed from a static standing pose. The setup, shown in Fig.~\ref{fig:setup}, uses a pendulum mechanism to release a mass from varying distances onto the robot in a repeatable manner. A cord allows retrieval of the weight post-impact, preventing interference during recovery.

The length $L$ of the pendulum and the height $H$ of the stand are respectively $1.28$m and $1.74$m and the mass $m$ of the weight is $0.9$kg. Assuming the pendulum mass is point-like, the kinetic energy of the mass at the moment of impact varies from $4$J to $7.3$J across the different tested configurations, which are chosen to produce impacts ranging from mild disturbances to guaranteed falls. During the experiments, the velocity of the mass is zero after impact, indicating a complete transfer of energy to the robot.
\vspace*{-0.5\baselineskip}
\vspace*{-0.1\baselineskip}

In this experiment, the robot is placed under the pendulum and subjected to repeated disturbances with varying impact intensities. After a disturbance, the robot waits to return to a stable position close to the target posture (cf. Fig. \ref{fig:dof}) before starting to walk in place.

The metric studied is the duration of disturbance rejection. Time measurement begins when any value in the state vector deviates by more than 5° from the neutral walking posture values. Once in this unstable state, all values in the state vector must have an error of less than 5° relative to the target posture for 0.5 seconds for the robot to be considered stable again and able to initiate walking. The time measurement ends at the beginning of the walking phase.

\section{Results}

% Métriques : Temps de relevage / retour à la stabilité après une pertubation (push and recovery) 
% Tableau de résultats comparant ancien relevage et une ou plusieurs policy sélectionnées pour des perturbations de force variable :
% - Temps de retour à la position stable
% - Nombre de chute (définition : un point de contact autre que les 2 pieds avec le sol)

% DISCUSSION (à mettre dans une section peut-être)

\subsection{Qualitative results}

\begin{figure}[tp]
    \centering
    \includegraphics[width=.48\textwidth]{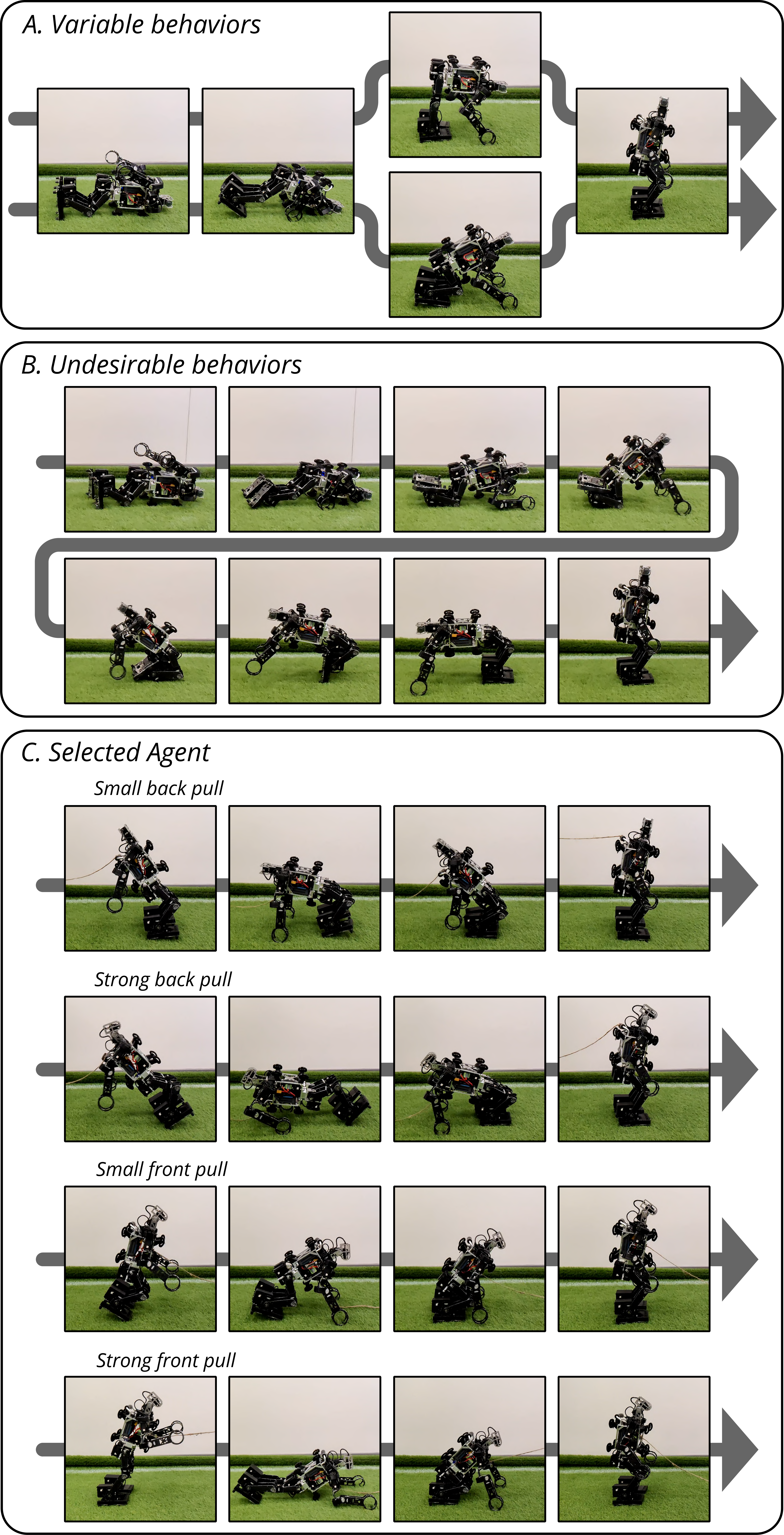}
    \vspace*{-1.2\baselineskip}
    \caption{Agents behaviors on real robot}
    \label{fig:qualitative}
    \vspace*{-1.5\baselineskip}
\end{figure}

The rapid training of the agents allows for multiple candidates to be trained to select the optimal FRASA candidate. Despite comparable training times and rewards, the behaviors that emerged are sometimes different. Fig.~\ref{fig:qualitative}.A shows an example of variation in the forward stand up task, where the robot either chose to squat or to remain with legs extended during stand up.

% \begin{figure}[h!]
%     \centering
%     \includegraphics[width=.4\textwidth]{img/Funky.pdf}
%     \caption{A TITRER}
%     \label{fig:funky}
% \end{figure}

The extensive agent environment exploration sometimes results in highly dynamic solutions, which proved feasible to implement on the real robot. One such solution is presented in Fig~\ref{fig:qualitative}.B. Although functional, the mechanical stresses and probable wear on the robots from performing this motion make it undesirable.

% \begin{figure}[h!]
%     \centering
%     \includegraphics[width=.4\textwidth]{img/Funky.pdf}
%     \caption{A TITRER}
%     \label{fig:funky}
% \end{figure}

The emergence of such varied behaviors leads to the selective choice of a candidate for FRASA. The selected candidate, whose reaction to disturbances of different intensities is shown in Fig.~\ref{fig:qualitative}.C, is chosen based on the following criteria: speed of recovery, robustness of movements, mechanical safety, and versatility of behaviors.

% \begin{figure}[h!]
%     \centering
%     \includegraphics[width=.47\textwidth]{img/Pull.pdf}
%     \caption{A TITRER}
%     \label{fig:pull}
% \end{figure}

\subsection{Quantitative results}

The stand up experiment is conducted by performing the stand up task 20 times for each side and each method. The average stand up times are presented in Table \ref{tab:expSU}. We observe that FRASA outperforms the KFB method in both possible configurations: FRASA completes the stand up from the supine position in $53\%$ and from the prone position in $68\%$ of the time required by the KFB method. This corresponds to average gains of $2.38$ seconds and $1.02$ seconds, respectively.

\begin{table}[tp]
    \centering
    \caption{Comparison of average stand up times for FRASA and KFB method from prone and supine positions}
    \resizebox{.47\textwidth}{!}{%
    \begin{tabular}{
      lcccc
    }
    \toprule
     & \multicolumn{2}{c}{\textbf{Prone position}} & \multicolumn{2}{c}{\textbf{Supine position}}\\
    \cmidrule(lr){2-3} \cmidrule(lr){4-5}
    \multicolumn{1}{c}{\text{Method}} & \multicolumn{1}{c}{FRASA} & \multicolumn{1}{c}{KFB} & \multicolumn{1}{c}{FRASA} & \multicolumn{1}{c}{KFB} \\
    \midrule
    \multicolumn{1}{c}{\text{Average Time / s}} & \textbf{2.135} $\pm 0.042$ & $3.154 \pm 0.005$ & \textbf{2.678} $\pm 0.178$ & $5.06\pm 0.008$\\
    \bottomrule
    \end{tabular}%
    }
    \label{tab:expSU}
    \par
    \vspace{1.7ex}
    The standard deviation is represented using the notation $\pm$
\end{table}

The results of the fall recovery experiment are detailed in Table \ref{tab:expFR}. Each configuration of distance, method, and side is repeated 10 times, and the mean instability time is calculated. An estimate of the kinetic energy transferred at the moment of impact is calculated by considering the pendulum weight as a point mass. The weakest backwards disturbance does not cause any imbalance in the robot.

\begin{table}[tp]
    \centering
    \caption{Comparison of average instability durations of FRASA and the KFB method after disturbances}
    \resizebox{0.48\textwidth}{!}{%
    \begin{tabular}{lccccccc}
    \toprule
     & \multicolumn{3}{c}{\textbf{Frontwards Disturbance}} & & \multicolumn{2}{c}{\textbf{Backwards Disturbance}} \\
    \cmidrule(lr){2-4} \cmidrule(lr){6-7}
    d / m & 0.56 & 0.75 & 0.89 & & 0.75 & 0.89 \\
    Energy / J & 4.0 & 5.5 & 7.3 & & 5.5 & 7.3 \\
    \midrule
    FRASA  / s & $0.54\pm 0.02$ & \textbf{2.41} $\pm 0.04$ & \textbf{2.44} $\pm 0.03$ & & $0.62\pm 0.10$ & \textbf{2.26} $\pm 0.11$ \\
    KFB / s & $0.57\pm 0.02$ & $5.96\pm 0.11$ & $5.74\pm 0.05$ & & $0.49\pm 0.10$ & $4.47\pm 0.23$ \\
    \bottomrule
    \end{tabular}%
    }
    \label{tab:expFR}
    \par
    \vspace{1ex}
    The standard deviation is represented using the notation $\pm$
    \vspace*{-1.8\baselineskip}
\end{table}

In all configurations, FRASA achieves comparable or shorter instability times than the KFB method. The only configuration where the KFB method has a lower mean instability time than FRASA is for a $5.5$J disturbance from the back. However, the standard deviation ranges for the two methods overlap, indicating that the difference is not statistically significant.

For the most significant impacts ($d=0.89$m representing $7.3$J), FRASA is able to reject the disturbance in $42\%$ of the time required by the KFB method for a front impact and in $51\%$ of the time for a back impact. FRASA also demonstrates superior performance for $5.5$J frontwards disturbance, where the return to stability takes only $40\%$ of the time compared to the KFB method.

Overall, FRASA demonstrates its superiority both in rejecting significant disturbances and in recovering from a prone or a supine position, surpassing the KFB method.

\section{Conclusion}

Experiments on the Sigmaban platform demonstrate FRASA's superiority over the KFB method in both standing up and fall recovery, achieving these tasks with increased efficiency. The integration of the CrossQ algorithm and the use of the robot's symmetry result in training times of around 30 minutes, which, to our knowledge, is unprecedented for successfully transferable humanoid stand up tasks.

The end-to-end nature of FRASA allows it to adapt to various disturbance intensities without requiring expert tuning. However, while leveraging the humanoid robot's symmetry accelerates training, it also limits the versatility of emerging behaviors, particularly in handling lateral disturbances.

To further enhance FRASA's performance, a promising direction would be to finalize its training directly on the robot (online) to potentially reduce the sim-to-real gap. Additionally, improving the modeling of actuators in our simulator could help bridge this gap even further.

\bibliographystyle{ieeetr}
\bibliography{biblio}

\end{document}